\documentclass[conference, letterpaper]{IEEEtran}

\IEEEoverridecommandlockouts
\usepackage{cite}
\usepackage{amsmath,amssymb,amsfonts}
\usepackage{algorithmic}
\usepackage{graphicx}
\usepackage{textcomp}
\usepackage{xcolor}
\usepackage{booktabs}
\usepackage{siunitx}
\usepackage[colorlinks=true, linkcolor=blue, citecolor=blue, urlcolor=blue]{hyperref}

\def\BibTeX{{\rm B\kern-.05em{\sc i\kern-.025em b}\kern-.08em
    T\kern-.1667em\lower.7ex\hbox{E}\kern-.125emX}}

\begin{document}

\title{FCBV-Net: Category-Level Robotic Garment Smoothing via Feature-Conditioned Bimanual Value Prediction\\
}

\author{\IEEEauthorblockN{Mohammed Daba}
\IEEEauthorblockA{\textit{School of Mechanical and Electrical Engineering} \\
\textit{University of Electronic Science and Technology of China}\\
Chengdu, Sichuan Province, P.R.of China \\
daba@std.uestc.edu.cn}
\and
\IEEEauthorblockN{Jing Qiu}
\IEEEauthorblockA{\textit{School of Mechanical and Electrical Engineering} \\
\textit{University of Electronic Science and Technology of China}\\
Chengdu, Sichuan Province, P.R.of China \\
qiujing@uestc.edu.cn}

}
\maketitle

\begin{abstract}
Category-level generalization for robotic garment manipulation, such as bimanual smoothing, remains a significant hurdle due to high dimensionality, complex dynamics, and intra-category variations. Current approaches often struggle, either overfitting with concurrently learned visual features for a specific instance or, despite Category-level perceptual generalization, failing to predict the value of synergistic bimanual actions. We propose the Feature-Conditioned bimanual Value Network (FCBV-Net), operating on 3D point clouds to specifically enhance category-level policy generalization for garment smoothing. FCBV-Net conditions bimanual action value prediction on pre-trained, frozen dense geometric features, ensuring robustness to intra-category garment variations. Trainable downstream components then learn a task-specific policy using these static features. In simulated PyFlex environments using the CLOTH3D dataset, FCBV-Net demonstrated superior category-level generalization. It exhibited only an 11.5\% efficiency drop (Steps80) on unseen garments compared to 96.2\% for a 2D image-based baseline, and achieved 89\% final coverage, outperforming an 83\% coverage from a 3D correspondence-based baseline that uses identical per-point geometric features but a fixed primitive. These results highlight that the decoupling of geometric understanding from bimanual action value learning enables better category-level generalization. Code, videos, and supplementary materials are available at the project website: \url{https://dabaspark.github.io/fcbvnet/}.

\end{abstract}


\section{Introduction}
\begin{figure}
\centering
\includegraphics[width=\columnwidth]{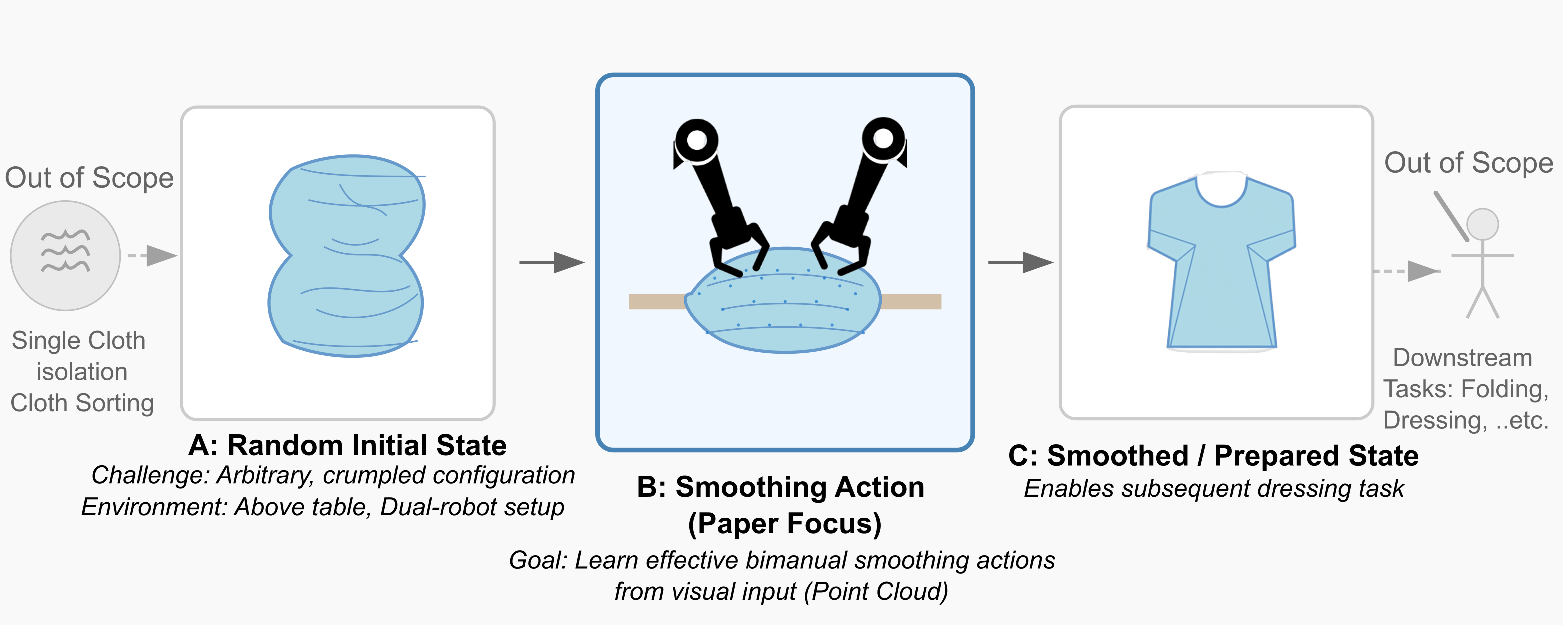}
\caption{Garment Smoothing is a crucial step in many downstream garment manipulation tasks (e.g., folding, dressing, etc.). It involves transforming a garment from a random state into a known, structured configuration.}
\label{fig_scope}
\end{figure}

Robotic manipulation of garments, crucial for applications such as assistive care and automated dressing \cite{r1, r2}, presents significant challenges such as high deformability, complex dynamics, and near-infinite dimensional configuration spaces \cite{r4}. Furthermore, items within the same category can display substantial intra-category variations in shape, size, and material, making robust manipulation a difficult task \cite{r5}. A key step in many garment handling pipelines, such as preparing a shirt for a person to wear, involves transforming an arbitrarily crumpled garment into a smoothed, predictable state (see Fig.~\ref{fig_scope}).

The core problem this paper addresses is enabling robots to perform bimanual garment smoothing efficiently across a specific garment category, not just on specific instances seen during training, but with strong generalization to \textit{unseen garments within the same category}. Existing approaches often struggle with this generalization. Methods that learn action-value functions directly can optimize bimanual coordination but may rely on learned state representations that overfit to training instances \cite{r6, r7, r8}. Conversely, while correspondence-based methods have good category-level generalization \cite{r5}, they typically use policy transfer and do not directly predict the \textit{outcome quality} of executing a specific bimanual action primitive.

To address these limitations, we propose the \textbf{Feature-Conditioned Bimanual Value Network (FCBV-Net)}. FCBV-Net operates directly on 3D point cloud and introduces a novel approach to bimanual action value prediction. 

The central hypothesis is that explicitly \textit{conditioning an action's value on robust, pre-trained per-point geometric features can dramatically improve generalization. We use dense features that are pre-trained for structural understanding and then keep them frozen during policy learning. This strategy decouples geometric understanding from learning task-specific interaction values.} As a result, the system can effectively learn to anticipate synergistic outcomes and generalize across unseen garment geometries within a category.

\begin{figure*}[!t]
\centering           
\includegraphics[width=\textwidth]{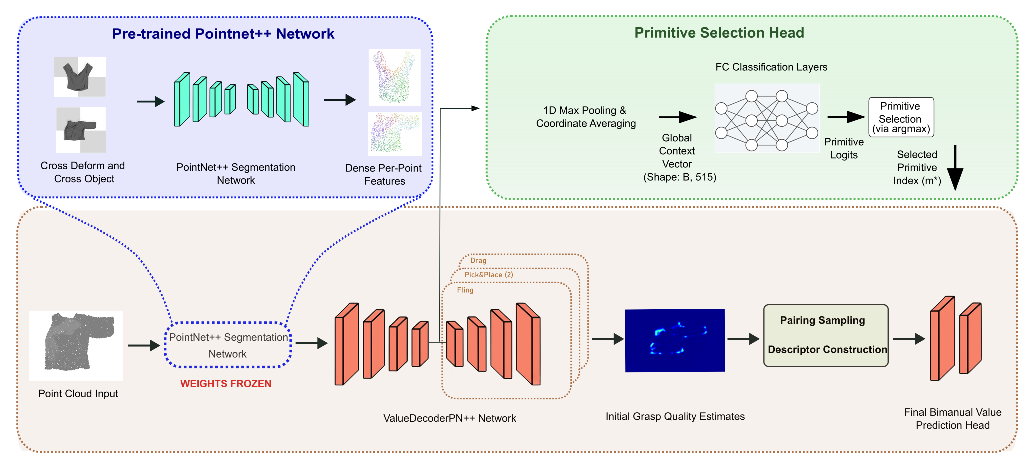}
\caption{Overview of the FCBV-Net Architecture. 
Dense geometric features (\(\mathbf{f}_p\)) are extracted by a pre-trained and frozen PointNet++ backbone (top-left). 
The main network (bottom) processes an input point cloud: a ValueDecoderPN++ Network, conditioned by these frozen features, predicts initial grasp quality estimates and embeddings. 
A parallel Primitive Selection Head (top-right), using global set abstraction and classification layers, determines the manipulation primitive. 
Finally, descriptors constructed from these outputs are evaluated by a Final Bimanual Value Prediction Head to yield the conditioned action value (\(Q_{\text{FCBV}}\)).}
\label{fig_main}
\end{figure*}

This strategy aims to decouple the learning of fundamental geometric understanding from the learning of task-specific, interaction-aware action values.

The main contributions of this work are:
\begin{itemize}
    \item The proposal of FCBV-Net, a novel architecture for category-level bimanual garment smoothing that conditions value prediction for actions using pre-trained, frozen per-point geometric features.
    \item A demonstration, through simulation in the PyFlex environment using the CLOTH3D dataset, that FCBV-Net achieves superior generalization performance on unseen `tops' category garments compared to a state-of-the-art 2D image-based baseline and a 3D correspondence-based policy transfer method.
\end{itemize}

\section{RELATED WORK}
\label{sec:related_work}

\subsection{Garment State Representation}
Early approaches often used 2D image features like corners and wrinkles~\cite{r9, r10}. The adoption of 3D point clouds provided richer geometric input~\cite{r5, r8}.
Beyond raw data, structured representations like skeletons or keypoints capture garment topology~\cite{r11, r12}, though sparse features may lack detail for precise control.
Consequently, dense representations, including learned visual correspondences~\cite{r13, r14} often via self-supervised contrastive learning~\cite{r5}, or dense visual affordances~\cite{r13}, have become prominent.
To leverage the detail of dense representations while ensuring generalization, FCBV-Net utilizes 3D point clouds processed into pre-trained, frozen per-point geometric features, providing a robust, deformation-invariant foundation that is crucial for informing downstream value prediction across varied garment instances.

\subsection{Learning Manipulation Policies}
Early methods for garment smoothing included heuristics~\cite{r15, r16} or goal-conditioned learning~\cite{r17}. Learning from Demonstration (LfD) frequently used to learn policies~\cite{r19}, Yet, these techniques necessitate several interactions.
Learning action-value functions (Q-functions) using self-supervised learning are increasingly applied to unfolding and smoothing~\cite{r21, r6}. These may predict value maps for single-point~\cite{r21} or bimanual interactions~\cite{r6, r8}, with training combining supervised signals and self-exploration.
To manage large action spaces, strategies include ranking sparse keypoint candidates~\cite{r8} or learning pair correspondence by sampling~\cite{r6}.
To address the risk of overfitting while still optimizing for complex interactions, FCBV-Net learns a Q-function conditioned on pre-trained features. This decoupling of geometry from policy learning allows the network to find optimal synergistic actions without being confined to fixed primitives or overfitting to training instances.

\subsection{Category-Level Generalization}
Many prior efforts were instance or task-specific~\cite{r6, r7}, with limited transferability. As generalization in this case is challenging due to intra-category variations in shape, size, material, and topology.
Strategies for generalization include learning dense visual correspondences~\cite{r19}, often via self-supervision on 3D model datasets~\cite{r5}, and generalizable structural representations like skeletal models~\cite{r11}.
FCBV-Net targets improved category-level generalization by conditioning learned action-values on robust, pre-trained dense visual correspondences features.

\subsection{Simulation for Garment Manipulation}
Simulation enables large-scale data collection for policy learning~\cite{r19, r21, r23}.
While general simulators like PyBullet~\cite{r24} have been adapted, specialized deformable object simulators like SoftGym~\cite{r25}, using NVIDIA FleX~\cite{r26} under the hood are common. However, these engines are frequently exhibiting a significant sim-to-real gap as a result of the lack of thorough sim-to-real algorithm designs.

Fast-computing simulators like PyFlex~\cite{r28} employ Position-Based Dynamics (PBD) to efficiently simulate deformable materials and generate realistic 3D observations. To ensure high-fidelity physical interactions and support large-scale data collection, our work is implemented and validated entirely within a customized PyFlex simulation ecosystem, augmented with a resource-aware multiprocessing manager and a robust simulation bridge.

\section{PROBLEM STATEMENT}
\label{sec:problem_statement}

Let the state of a garment at time $t$ be represented by a 3D point cloud observation $\mathcal{O}_t \in \mathbb{R}^{N \times 3}$, where $N$ is the number of points. 
The primary objective is to determine a sequence of actions $\{a_0, a_1, \ldots, a_{k-1}\}$ to transit the garment from an arbitrary initial configuration $\mathcal{O}_0$ to a smoothed, flattened state $\mathcal{O}_k$ that maximizes our objective reward, see equation~\eqref{eq:reward}. 
This task requires generalization across different instances within a specific category, denoted by $\mathcal{G}$, even if those instances $\mathcal{G}_{\text{unseen}} \notin \mathcal{G}_{\text{train}}$ were not encountered during training.

Each bimanual action $a_t$ is defined by a tuple:
\begin{equation}
a_t = (m_t, p_{t,1}, p_{t,2})
\label{eq:action_definition}
\end{equation}
where $m_t \in \mathcal{M}$ is a manipulation primitive selected from a discrete set $\mathcal{M} = \{\text{Fling, Drag, PickPlace, Done}\}$. 
The term $p_{t,i} \in \mathbb{R}^{2}$ represents the continuous 2D grasp point coordinates on the projected image plane. Because the simulation evaluates physical actions, these 2D coordinates are subsequently raycasted back to 3D for execution. Note that while physical deployments require planar grasp orientations, we intentionally abstract rotation in this formal action space since the current simulation environment utilizes idealized, point-based particle attachment for the grippers.

The core challenge is to train a network that learns a policy $\pi(a_t | \mathcal{O}_t)$ that selects an optimal action $a_t^*$ to maximize our objective. 
This policy must be robust to significant intra-category variations, and deformations. To achieve this, we pass the input through a frozen, dense geometric feature network, resulting in a feature vector $\mathbf{f}_p \in \mathbb{R}^{D}$ extracted for each point $p \in \mathcal{O}_t$
These features are pre-trained to be invariant to deformation and garment instance within a category~\cite{r5}.

The problem is to effectively leverage pre-trained frozen features $\mathbf{f}_p$ to predict $a_t$, improving category-level generalization as measured by the metrics defined in~\ref{subsec:evaluation_metrics}.

\section{METHOD}
\label{sec:method}

Our approach learns to predict the value of bimanual actions by leveraging geometric understanding that is robust of deformation and intra-category variations. The architecture, depicted in Fig.~\ref{fig_main}, processes an input point cloud $\mathcal{O}_t$ to select an optimal bimanual action $a_t^*$.

\subsection{Dense Geometric Feature Extraction}
\label{subsec:feature_extraction}

Taking inspiration from~\cite{r5}, a dense geometric feature extractor, $F_{\text{feat}}$, implemented as a PointNet++~\cite{r30} segmentation version. Given an input point cloud $\mathcal{O}_t \in \mathbb{R}^{N \times 3}$, $F_{\text{feat}}$ outputs per-point dense feature vectors $\mathbf{F}_{\mathcal{O}_t} = \{\mathbf{f}_p \in \mathbb{R}^{D} | p \in \mathcal{O}_t\}$, where $D=512$ is the feature dimensionality. $F_{\text{feat}}$ is pre-trained using a self-supervised contrastive learning strategy using self-play and skeletons~\cite{r11} on 'tops' category garments from CLOTH3D~\cite{r31}, employing cross-deformation (LCD) and cross-object (LCO) consistency losses as in~\cite{r5}. Specifically, LCD loss is formulated as:
\begin{equation}
\mathcal{L}_{\text{LCD}} = -\log \frac{\exp(\text{sim}(\mathbf{f}_p, \mathbf{f}_{p'}) / \tau)}{\sum_{j=1}^{M} \exp(\text{sim}(\mathbf{f}_p, \mathbf{f}_{p_j}) / \tau)}
\label{eq:lcd_loss}
\end{equation}
where $\mathbf{f}_p$ and $\mathbf{f}_{p'}$ are feature vectors of corresponding points on different deformations of the same garment, $\text{sim}(\cdot, \cdot)$ denotes the dot product as a measure of two feature vectors similarity, $\tau$ is a temperature parameter, and the sum is over $M-1$ negative samples. A similar formulation, $\mathcal{L}_{\text{LCO}}$, is used for corresponding points on different garment instances. After pre-training, the weights of $F_{\text{feat}}$ are frozen, providing a generalizable geometric representation for all downstream components. The input to subsequent trainable modules processes the point cloud geometry and features separately. The geometric coordinates $\mathcal{O}_{t}$ are used for spatial abstraction, while the feature channels $\mathbf{X}_{\text{in}}$ differ from the geometry. $\mathbf{X}_{\text{in}}$ is formed by concatenating the normalized RGB values from the sensor with the frozen, normalized per-point geometric features $\mathbf{F}_{\mathcal{O}_t}$, resulting in an input dimensionality of $\mathbb{R}^{N \times (3+D)}$.

\begin{figure}[htbp]
    \centering
    \includegraphics[width=\columnwidth]{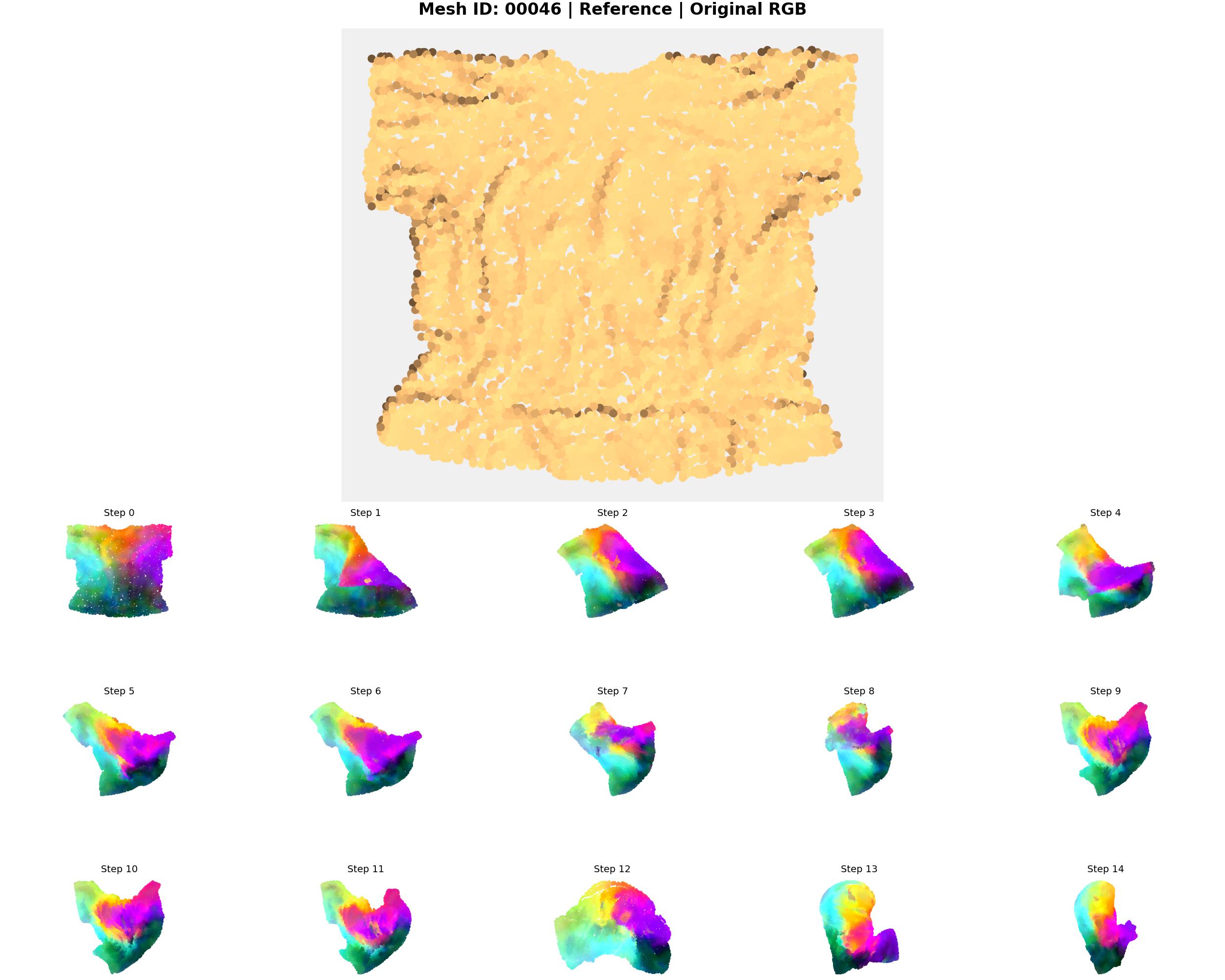} 
    \caption{PCA-projected visualization of the pre-trained PointNet++ features. The consistency of color patterns across different deformation steps demonstrates that the extracted features successfully capture intrinsic geometric topology invariant to the garment's spatial configuration.}
    \label{fig:feature_invariance}
\end{figure}

As illustrated in Fig.~\ref{fig:feature_invariance}, projecting these high-dimensional features into RGB space via Principal Component Analysis (PCA) reveals highly consistent semantic coloring across severe deformations, visually confirming the robustness of the frozen geometric representation.

\begin{table*}[t]
\caption{System-level performance of FCBV-Net and baselines on the bimanual garment smoothing task, evaluated on seen and unseen `tops' from the CLOTH3D dataset. Results are averaged over 20 trials.}
\label{tab:quantitative_results}
\centering
\begin{tabular}{l S[table-format=1.1] S[table-format=1.2] S[table-format=1.1] S[table-format=1.2] S[table-format=2.1]}
\toprule
\textbf{Method} & \multicolumn{2}{c}{\textbf{Seen Garments}} & \multicolumn{2}{c}{\textbf{Unseen Garments}} & {\textbf{Generalization}} \\
\cmidrule(lr){2-3} \cmidrule(lr){4-5}
& {Steps80(\boldmath$\downarrow$)} & {FinalCovH5(\boldmath$\uparrow$)} & {Steps80(\boldmath$\downarrow$)} & {FinalCovH5(\boldmath$\uparrow$)} & {Drop80(\%)(\boldmath$\downarrow$)} \\
\midrule
FCBV-Net (Ours)    & \textbf{2.6} & \textbf{0.91} & \textbf{2.9} & \textbf{0.89} & 11.5 \\
Sim-SF             & \textbf{2.6} & 0.89          & 5.1          & 0.79          & 96.2 \\
UGM-PolicyTransfer & 2.8          & 0.84          & 3.0          & 0.83          & \textbf{7.1} \\
\bottomrule
\end{tabular}
\end{table*}

\subsection{Action Proposal and Value Prediction}
\label{subsec:action_value_prediction}

FCBV-Net employs several trainable modules to propose and evaluate candidate bimanual actions. 

The \textbf{Initial Value Prediction and Embedding Network (ValueDecoderPN++)} is a modified version of the segmentation variant of PointNet++~\cite{r30} based architecture. It takes the point cloud geometry and the concatenated features $\mathbf{X}_{\text{in}}$ (RGB + Frozen Geometric Features) as input. The network comprises a shared encoder-decoder backbone and specialized task-specific heads. The shared backbone utilizes Set Abstraction (SA) modules to encode global geometry into latent features $\mathbf{Z}_{\text{enc}} \in \mathbb{R}^{N \times D'}$, followed by shared Feature Propagation (FP) modules that upsample features back to the original points. 
The propagated features are then fed into parallel, specialized multi-layer perceptron (MLP) branches implemented via 1D convolutions. Specifically, for the distinct manipulation primitives (\textit{Fling}, \textit{Pick\&Place}, and \textit{Drag}), the network utilizes four separate decoders (one for Fling, two for Pick\&Place, and one for Drag) to capture their specific geometric requirements. Each head outputs a per-point initial unconditioned quality estimate $Q_{\text{unc-cond}}(p) \in \mathbb{R}$ and a learned point embedding $\mathbf{e}(p) \in \mathbb{R}^{M_e}$, where $M_e=8$. These are grouped as $\mathbf{Y}_{\text{val}}^m(p) = [Q_{\text{unc-cond}}(p), \mathbf{e}(p)]$. The network focuses on identifying optimal contact points on the garment surface.

\textbf{The Primitive Selection Head (Head\textsubscript{m})} module takes the downsampled features and coordinates from the final Set Abstraction module of the ValueDecoderPN++ encoder. It aggregates this bottleneck output into a single global context vector $\mathbf{z}_{\text{global}}$ by concatenating the spatial mean of the point coordinates with the max-pooled features. Head\textsubscript{m} then processes this vector through fully connected layers to predict a probability distribution $P(m | \mathbf{z}_{\text{global}})$ over the manipulation primitives $m \in \mathcal{M}$. The primitive $m^*$ with the highest probability is selected.

\textbf{For Candidate Action Sampling and Descriptor Construction}, given a selected primitive $m$, candidate grasp points $p_1$ and $p_2$ are sampled from the predicted unconditioned value map $Q_{\text{unc-cond}}^{(m)}$. For each sampled candidate point $p_i$, a comprehensive correspondence descriptor vector $d_i$ is constructed:
\begin{equation}
d_i = \left[Q_{\text{unc-cond}}^{(m)}(p_i) \parallel u_{\text{norm},i} \parallel v_{\text{norm},i} \parallel \mathbf{e}^{(m)}(p_i)\right]^T
\label{eq:descriptor}
\end{equation}
where $\parallel$ denotes vector concatenation, and $\mathbf{e}^{(m)}(p_i) \in \mathbb{R}^{D_e}$ is the learned point embedding. To ensure numerical stability for the downstream networks, the continuous pixel coordinates $(\mathbf{u}_{\text{norm}}, \mathbf{v}_{\text{norm}})$ projected from the 3D point $p_i$ are normalized to a $[-1,1]$ range relative to the image plane dimensions.

\textbf{The Final Conditioned Bimanual Value Head (NN\textsubscript{Q})} is a Multi-Layer Perceptron (MLP) that evaluates the synergistic quality of action pairs. It takes a concatenated vector $\mathbf{x} = [\mathbf{d}_{p_1} \parallel \mathbf{d}_{p_2}]^T$ as input. Given a learned point embedding dimension of $D_e=8$, the dimension of a single descriptor is $1 + 2 + 8 = 11$, resulting in a 22-dimensional input to NN\textsubscript{Q}. It outputs a scalar value $Q_{\text{FCBV}}(\mathcal{O}_t, a_t) \in \mathbb{R}$, which represents the expected reward~\eqref{eq:reward} of executing the synergistic bimanual action. During inference, a deterministic policy is employed, selecting the action pair $a_t^*$ that maximizes this predicted quality.

\begin{figure*}[thpb] 
      \centering
      \includegraphics[width=0.7\textwidth]{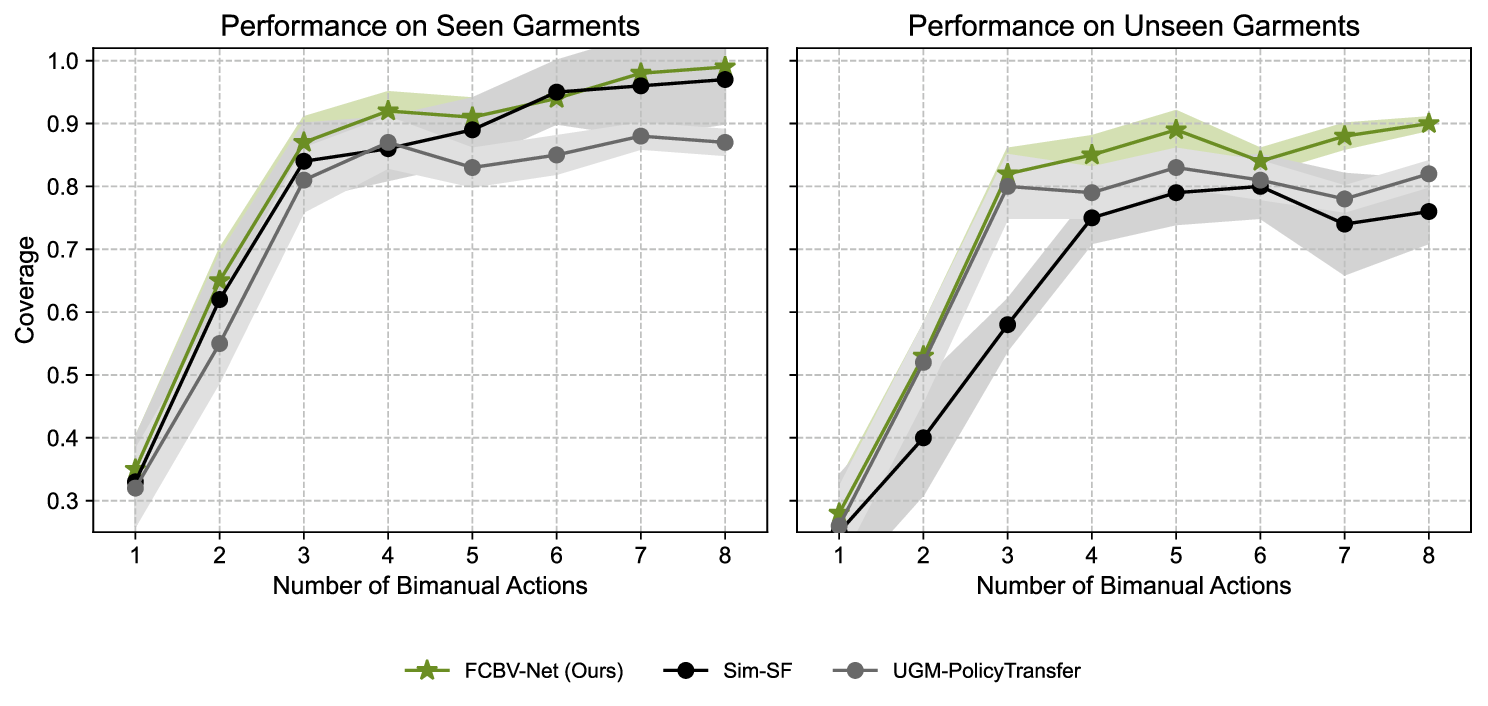} 
      \caption{Progression of normalized garment coverage over number of bimanual actions on seen garments (left) and unseen garments (right). FCBV-Net demonstrates robust performance and generalization, consistently achieving higher coverage compared to baselines.}
      \label{fig:coverage_progression}
\end{figure*}

\subsection{Training Procedure}
\label{subsec:training}

The trainable components ValueDecoderPN++, Head\textsubscript{m}, and NN\textsubscript{Q} are trained end-to-end iteratively, subsequent to the frozen $F_{\text{feat}}$. Training combines an initial dataset of human-annotated actions with data collected via self-supervised interaction in the simulation environment.

Taking inspiration from~\cite{r6}, the primary learning signal for training is the reward $r_t$, derived from the change in the visible surface area of the garment as it is smoothed. For a transition from state $\mathcal{O}_t$ to $\mathcal{O}_{t+1}$ after action $a_t$:
\begin{equation}
r_t = \max\big(\tanh\big[\alpha \cdot \Delta\text{Cov}(\mathcal{O}_t, \mathcal{O}_{t+1})\big], 0\big)
\label{eq:reward}
\end{equation}
where $\Delta\text{Cov}(\mathcal{O}_t, \mathcal{O}_{t+1})$ is the change in the normalized projected 2D coverage area of the garment calculated via the physics simulator. We set $\alpha=2.0$ to scale the coverage improvements effectively within the tanh non-linearity.

The total loss $\mathcal{L}_{\text{total}}$ is a weighted sum of several components:
\begin{equation}
\mathcal{L}_{\text{total}} = (\sum_{m \in \mathcal{M}_{\text{grasp}}} \lambda_h \mathcal{L}_{\text{heatmap}}^m) + \lambda_v \mathcal{L}_{\text{value}} + \lambda_c \mathcal{L}_{\text{class}}
\label{eq:total_loss}
\end{equation}
where $\mathcal{M}_{\text{grasp}}$ are primitives requiring heatmap prediction, and $\lambda_h, \lambda_v, \lambda_c$ are loss weighting coefficients.

The primitive-specific heatmap loss, $\mathcal{L}_{\text{heatmap}}^m$, trains the decoders $D_{\text{val}}^m$. Since the ground-truth annotations are provided as 2D heatmaps in the camera plane, we project the predicted 3D per-point qualities $Q_{\text{unc-cond}}(p)$ onto the 2D image plane using the camera intrinsics. Let $H^m_{\text{2D}}$ be the projected 2D predicted map and $T^m_{\text{GT}}$ be the ground-truth 2D Gaussian heatmap scaled by reward $r_t$. The loss is a dynamically weighted Binary Cross-Entropy (BCE):
\begin{equation}
\begin{split}
\mathcal{L}_{\text{heatmap}}^m &= \frac{1}{|\Omega|} 
\sum_{(u,v) \in \Omega} \Bigg[ W_{\text{dyn}}^m(u,v) \\
&\quad \cdot \text{BCE}\left(H^m_{\text{2D}}(u,v), r_t \cdot T^m_{\text{GT}}(u,v)\right) \Bigg]
\end{split}
\label{eq:heatmap_loss}
\end{equation}
where $\Omega$ represents the set of image pixels. The dynamic weight $W_{\text{dyn}}^m$ up-weights high-confidence regions to focus learning on peaks:
\begin{equation}
W_{\text{dyn}}^m(u,v) = w_{\text{min}} + H^m_{\text{2D}}(u,v) \cdot (w_{\text{max}} - w_{\text{min}})
\label{eq:dynamic_weight}
\end{equation}
This weighting scheme pushes the model to focus on correcting major errors and learning the exact locations of target peaks, rather than being overwhelmed by large, less informative regions in the heatmap.

The bimanual value loss, $\mathcal{L}_{\text{value}}$, trains the final bimanual value head NN\textsubscript{Q}. It uses a contrastive strategy with the BCEWithLogitsLoss function, denoted $\mathcal{L}_{\text{BCEWL}}$:
\begin{equation}
\mathcal{L}_{\text{value}} = \frac{1}{N_{\text{pairs}}} \sum_{i=1}^{N_{\text{pairs}}} \mathcal{L}_{\text{BCEWL}}(Q_{\text{FCBV}}(\mathbf{d}_{p_1}^i, \mathbf{d}_{p_2}^i), y^i)
\label{eq:value_loss}
\end{equation}
where $N_{\text{pairs}}$ is the number of constructed descriptor pairs per training sample. For the $i$-th pair $(\mathbf{d}_{p_1}^i, \mathbf{d}_{p_2}^i)$, the target $y^i = r_t$ if the pair corresponds to a ground-truth interaction, and $y^i = 0$ for negative or mixed pairs.

The primitive classification loss, $\mathcal{L}_{\text{class}}$, trains Head\textsubscript{m}. It is a weighted Cross-Entropy (CE) loss:
\begin{equation}
\mathcal{L}_{\text{class}} = w_{\text{annot}} \cdot \text{CE}(\text{logits}_{\text{class}}, \text{idx}(m^*), \mathbf{W}_{\text{class}})
\label{eq:class_loss}
\end{equation}
where $\text{logits}_{\text{class}}$ are raw outputs from Head$_m$, $\text{idx}(m^*)$ is the ground-truth primitive class index, and $\mathbf{W}_{\text{class}}$ is a static tensor with per-class weights to address data imbalance. The scalar $w_{\text{annot}}$ down-weights the loss for self-supervised data compared to human-annotated data. This loss is computed only for training samples where the observed reward $r_t$ exceeds threshold $r_c$.

\begin{figure*}[htbp] 
    \centering
    \includegraphics[width=0.9\textwidth]{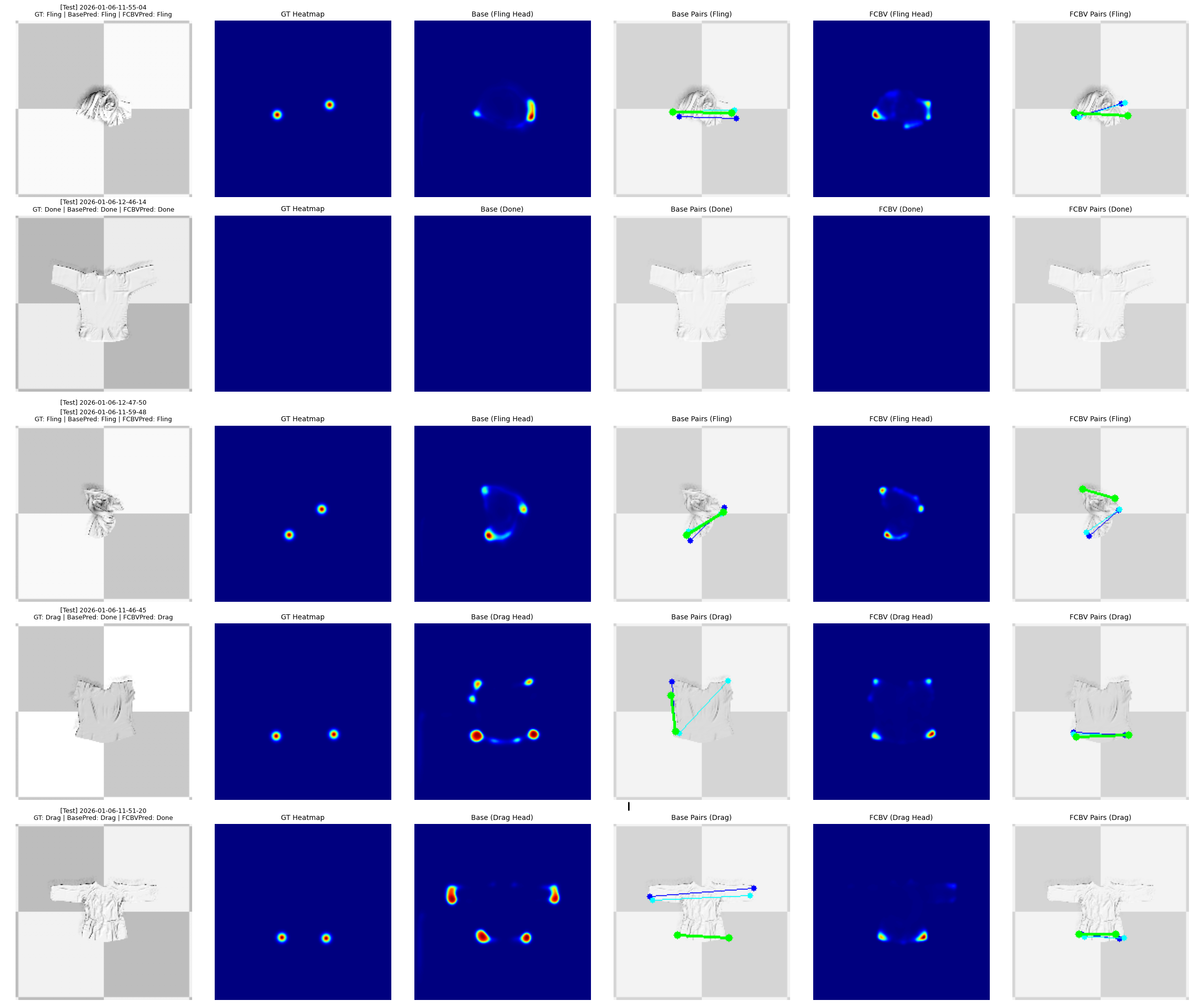} 
    \caption{Qualitative comparison of final bimanual action synthesis on unseen garment configurations. FCBV-Net (right) effectively isolates optimal functional points to predict parallel, synergistic action lines, actively avoiding the severe kinematic mismatches (e.g., crossed-arms) and asymmetric grasping common in the 2D baseline (middle).}
    \label{fig:action_synthesis}
\end{figure*}

\section{EXPERIMENT SETUP}
\label{sec:experiment_setup}

\subsection{Simulation Environment and Datasets}
\label{subsec:sim_env_datasets}

Experiments utilize a highly stable, customized simulation framework based on the \textbf{NVIDIA FleX} physics engine (PyFlex bindings)~\cite{r28}. To accommodate the complex physical interactions required for bimanual garment manipulation, this setup is encapsulated within a robust, containerized architecture featuring a custom simulation driver. It provides high-fidelity particle-based simulation for deformable objects and generates 3D point clouds ($\mathcal{O}_t \in \mathbb{R}^{N \times 3}$) from a simulated overhead depth camera, incorporating synchronized RGB-D rendering.

The ``tops'' category from the \textbf{CLOTH3D dataset}~\cite{r31} is used, featuring 499 instances such as t-shirts and hoodies. These instances are preprocessed and split into a \textbf{Training Set} of 450 instances and an \textbf{Unseen Set} of 49 instances. The training set is used for pre-training the frozen feature extractor and for all downstream FCBV-Net policy learning. From this set, we create a \textit{Deformation Training Pool} by subjecting each instance to simulated physical interactions, resulting in multiple crumpled configurations with up to 31 states per instance. Similarly the unseen set forms a \textit{Deformation Unseen-Instance Test Pool} and is never encountered during any training phase.

For policy learning, an aggregated dataset totaling 1300 action data points is constructed iteratively from the Deformation Training Pool. Initially, 200 human-annotated configurations are collected. Subsequently, the policy autonomously collects approximately 1100 self-supervised interaction trajectories. To enhance robustness and address primitive imbalances, 700 states from the self-supervised collection were re-annotated by human experts to provide corrective actions. This aggregated dataset of 1300 actions is then split: 1040 actions (80\%) form the training set, and 260 actions (20\%) form the validation set.

\subsection{Action Primitives}
\label{subsec:action_primitives}
The FCBV-Net policy selects bimanual actions from $\mathcal{M} = \{\text{Fling, Drag, PickPlace, Done}\}$. Execution logic follows established garment manipulation strategies~\cite{r6, r21}.

\subsection{Evaluation Metrics}
\label{subsec:evaluation_metrics}
Policy performance is quantified using the following metrics, primarily based on the normalized projected 2D workspace coverage of the garment relative to its fully flattened state:

\begin{itemize}
    \item \textbf{Steps80}: Average number of bimanual actions required to reach $\ge 80\%$ coverage, measuring efficiency.
    \item \textbf{FinalCovH5}: Average normalized coverage achieved after exactly five actions, assessing policy effectiveness under a strict action budget.
    \item \textbf{Drop80}: Performance degradation in Steps80 when transitioning from operating on familiar (seen) to novel (unseen) garment geometries, quantifying policy generalization.
\end{itemize}
Results are averaged over 20 trials per condition with a maximum of 8 actions per trial.

\begin{figure*}[htbp] 
    \centering
    \includegraphics[width=0.9\textwidth]{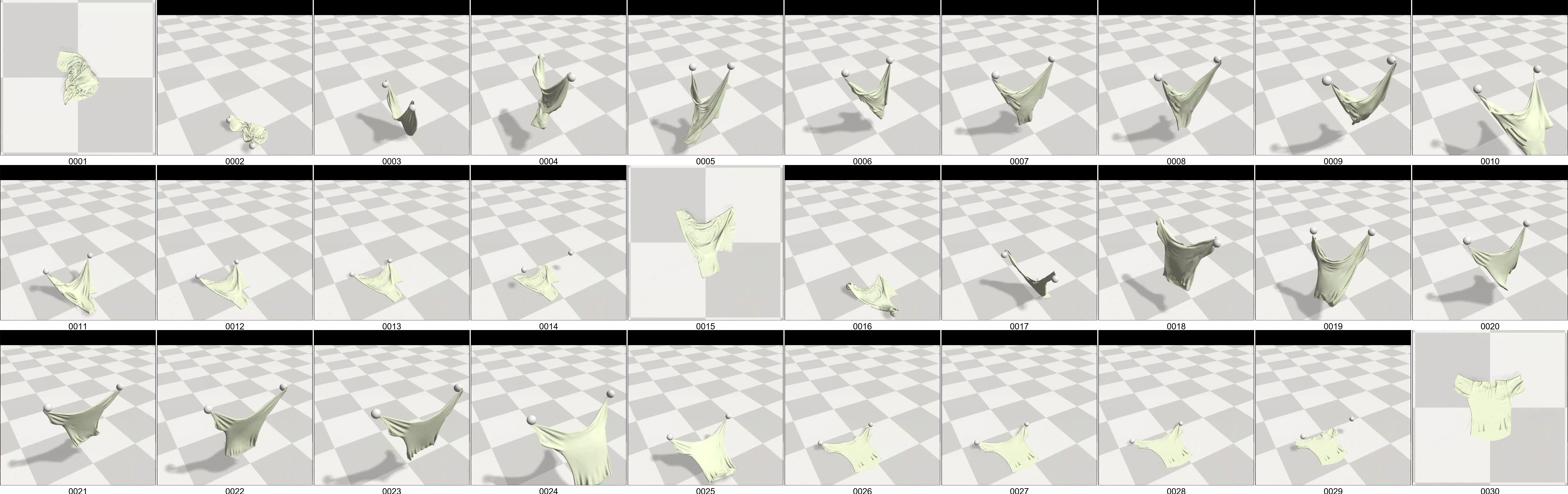} 
    \caption{Sequential visualization of an autonomous multi-step bimanual smoothing rollout executed by the FCBV-Net policy. The network dynamically adapts to the changing geometric state, successfully chaining primitives to completely flatten the garment from an unstructured initialization.}
    \label{fig:sequential_rollout}
\end{figure*}

\subsection{Baseline Methods}
\label{subsec:baselines}
To evaluate FCBV-Net's contributions, we compare it against two strong baselines:
\begin{enumerate}
    \item \textbf{Sim-SF}: A simulated version of a modified version of speedfolding~\cite{r6} method. It is a 2D image-based bimanual manipulation method (ResNeXt-50 encoder, U-Net decoders) where visual features are learned concurrently with the policy. Re-implemented and trained within PyFlex environment.
    \item \textbf{UGM-PolicyTransfer}: UniGarmentManip paper~\cite{r5} implementation that uses the \textit{identical frozen per-point geometric features ($F_{\text{feat}}$)} as FCBV-Net but employs a fixed `Fling' primitive policy based on geometric correspondence matching to template points, without learned value prediction.
\end{enumerate}

\subsection{Implementation Details}
\label{subsec:implementation}

The network processes input point clouds of $N=10,000$ points. The frozen geometric feature extractor, $F_{\text{feat}}$, outputs $D=512$ dimensional per-point features. The model operates at a resolution of $192 \times 192$ for the projected 2D auxiliary losses.

The training process is governed by several critical parameters. To ensure gradient stability, the target variable for the heatmap loss is directly scaled by the executed action's normalized task reward $r_t$, while the terminal `Done' classification utilizes a distinct learning signal. 
To enhance model robustness and prevent overfitting, data augmentation is applied on-the-fly via a continuous spatial transformation matrix. This applies synchronized transformations across modalities, utilizing a uniform scaling factor $s \sim \mathcal{U}(0.8, 1.2)$, a random rotation angle $\theta \sim \mathcal{U}(-\pi, \pi)$, and independent horizontal and vertical reflections with a Bernoulli probability $p=0.5$ to boost sample efficiency while maintaining geometric consistency.
The primitive classification loss, $L_{\text{class}}$, is only computed for samples where the observed reward exceeds a threshold of $r_c=0.3$. To prioritize learning from trusted labels, the loss contribution from self-supervised data is down-weighted by a factor of 0.01 compared to human-annotated data. 
The total loss weighting coefficients in Eq.~\eqref{eq:total_loss} are set to $\lambda_c = 80.0$ and $\lambda_v = 2500.0$.
The dynamic weights for the heatmap loss, $W_{\text{dyn}}$, are interpolated between $w_{\text{min}}$ and $w_{\text{max}}$. These bounds are source-dependent: high-confidence human annotations utilize a range of $[0.5, 1.0]$ to enforce stricter spatial precision, whereas self-supervised rollouts utilize a relaxed range of $[0.04, 1.0]$ to account for potential noise in the autonomously collected labels. The downstream FCBV-Net components are trained over 100 epochs using a batch size of 8. All trainable components are optimized end-to-end using the Adam optimizer~\cite{adam2014method} with an initial learning rate of 4e-4, a weight decay of 1e-6, and an \texttt{ExponentialLR} scheduler with a decay rate ($\gamma$) of 0.97.

\section{RESULTS AND ANALYSIS}
\label{sec:results_analysis}

We evaluated FCBV-Net against Sim-SF and UGM-PolicyTransfer on seen and unseen `tops' from CLOTH3D. The quantitative results are summarized in Table~\ref{tab:quantitative_results} and the progression of garment coverage is depicted in Fig.~\ref{fig:coverage_progression}.

On \textbf{seen garments}, FCBV-Net effectively learns smoothing policies, requiring an average of 2.6 steps to reach 80\% coverage. As shown in Fig.~\ref{fig:coverage_progression}, it rapidly surpasses 90\% coverage. Sim-SF matches this efficiency with 2.6 Steps80 but achieves a lower final coverage of 0.89. The UGM-PolicyTransfer baseline is less efficient at 2.8 Steps80 and also attains a lower final coverage of 0.84, highlighting the advantage of FCBV-Net's learned, adaptive policy over a fixed heuristic even on familiar items.

The critical evaluation is on \textbf{unseen garments}, where FCBV-Net demonstrates strong category-level generalization. It maintains high performance with 2.9 Steps80 and 0.89 FinalCovH5. This minimal efficiency decline results in a Generalization Drop of only 11.5\%, confirming its robust adaptation to novel geometries. As shown in Fig.~\ref{fig:coverage_progression}, FCBV-Net consistently achieves the highest coverage on unseen garments.

In contrast, the 2D image-based Sim-SF exhibits significant performance degradation. Its efficiency plummets, with Steps80 increasing from 2.6 to 5.1, resulting in a 96.2\% Generalization Drop. Its final coverage also falls to 0.79. This poor performance suggests its concurrently learned 2D visual features overfit to training instances, failing to generalize.

UGM-PolicyTransfer, leveraging the same pre-trained 3D features as FCBV-Net, shows excellent generalization in terms of efficiency, with its Steps80 increasing from 2.8 to 3.0. This result validates the category-level robustness of the underlying per-point geometric features. In contrast, FCBV-Net achieves a higher final coverage of 0.89, highlighting the effectiveness of its learned, multi-primitive policy over a fixed heuristic. While its feature-based Fling heuristic generalizes well, it is ultimately less effective than FCBV-Net's multi-primitive policy at achieving smoother state.

Our core strategy is to decouple robust geometric understanding from the learning of a task-specific value function. This approach allows FCBV-Net to outperform key baselines. It avoids the overfitting common in 2D methods and achieves a more complete smoothed state than a 3D correspondence-based method that relies on a fixed primitive.

The quantitative superiority of FCBV-Net on unseen garments is directly tied to its precision in spatial reasoning and bimanual synthesis. As illustrated in Fig.~\ref{fig:action_synthesis}, the 2D image-based baseline frequently struggles with depth ambiguities and perspective distortions on novel topologies. This results in diffuse activation heatmaps and the synthesis of physically flawed bimanual pairs, such as structurally asymmetric grasps or ``crossed-arms'' kinematic collisions. By explicitly conditioning the value prediction on deformation-invariant 3D geometric features, FCBV-Net effectively isolates optimal functional regions. Consequently, the final bimanual value head reliably synthesizes parallel, cohesive grasp pairs that strictly align with the intended manipulation strategy, maintaining high confidence even on entirely unseen garment instances.

Beyond single-step predictions, the robustness of the FCBV-Net policy is validated by its ability to execute closed-loop, multi-step manipulation trajectories. Fig.~\ref{fig:sequential_rollout} depicts an autonomous rollout of the FCBV-Net policy interacting with a highly crumpled, unseen garment. By continuously re-evaluating the updated 3D geometric state after each physical interaction, the network successfully chains a sequence of synergistic primitives. This demonstrates the policy's capacity to systematically resolve complex folds and achieve a fully prepared, smoothed state without falling into sub-optimal action cycles.

\begin{figure}[htbp]
    \centering
    \includegraphics[width=0.95\columnwidth]{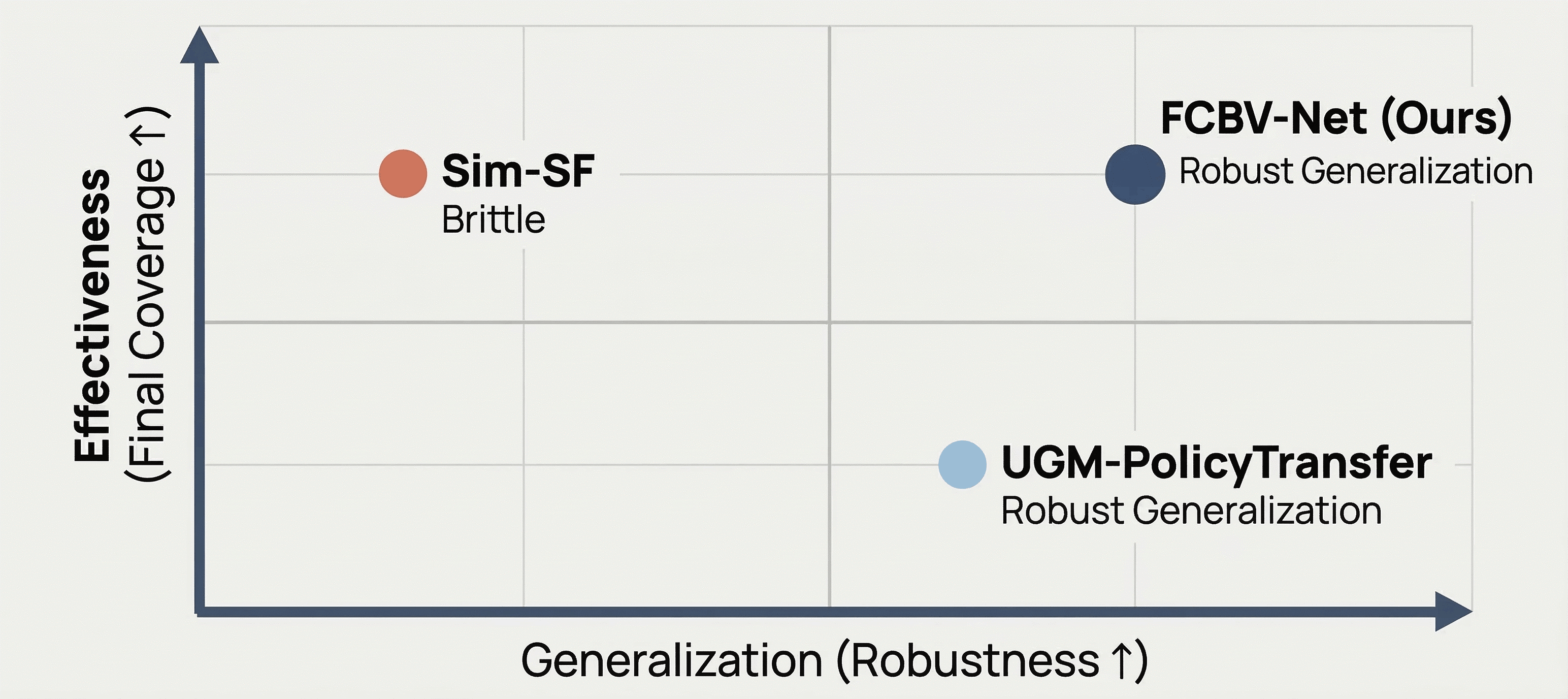} 
    \caption{Conceptual Performance Quadrant demonstrating the impact of the decoupling hypothesis. FCBV-Net overcomes the traditional trade-off, achieving both the high manipulation effectiveness characteristic of learned policies and the robust category-level generalization of geometric correspondence methods.}
    \label{fig:performance_quadrant}
\end{figure}

Ultimately, these empirical results validate the core decoupling hypothesis proposed in this work. As summarized in the performance quadrant in Fig.~\ref{fig:performance_quadrant}, existing paradigms face a strict dichotomy: concurrent learning approaches (Sim-SF) achieve high task effectiveness on familiar items but are highly brittle to novel geometries, whereas fixed-heuristic transfer methods (UGM-PolicyTransfer) generalize well but sacrifice absolute task effectiveness. FCBV-Net successfully breaks this trade-off. By explicitly decoupling robust 3D geometric understanding from task-specific bimanual value prediction, the proposed architecture achieves both state-of-the-art final manipulation effectiveness and robust category-level generalization.

\section{CONCLUSION}
\label{sec:conclusion}

This paper introduced the Feature-Conditioned Bimanual Value Network (FCBV-Net) to address the challenge of category-level generalization in garment smoothing. In simulation, FCBV-Net demonstrated superior generalization on unseen garments compared to relevant baselines. It maintained high efficiency with only an 11.5\% performance drop, unlike a 2D image-based approach that degraded by 96.2\%. Compared to a 3D correspondence-based method using identical features but a fixed `Fling` primitive, FCBV-Net achieved a higher final garment coverage (89\% vs. 83\%) on unseen items. This shows the effectiveness of our method in achieving a more smoothed state. The results validate that our method of leveraging robust, pre-trained 3D features enables a learned policy to generalize effectively across novel garment instances.

Despite these promising simulation results, a primary limitation of this study is the lack of real-world experiments. The current evaluation relies entirely on the customized PyFlex physics simulator, which inherently abstracts away several physical complexities. Bridging the sim-to-real gap stands as a critical key area for future work. Successful physical robot deployment will require demonstrating the policy's robustness to real-world sensor noise, severe depth camera occlusions, and the unpredictable friction and continuum dynamics of actual fabrics. Furthermore, to mitigate execution inaccuracies inherent to physical robot grippers and open-loop heuristics, future research must explore the integration of multi-modal sensory feedback (e.g., tactile or force-torque sensors) to enable reactive, closed-loop manipulation.

\bibliographystyle{IEEEtran}
\bibliography{IEEEabrv,bibfile}

\end{document}